\documentclass[conference]{IEEEtran}
\usepackage{cite}

\usepackage[utf8]{inputenc}
\usepackage[ruled,vlined]{algorithm2e}
\usepackage{amsmath,amssymb,amsfonts}
\usepackage{tabularx}
\usepackage{multirow}
\usepackage{graphicx}
\usepackage{balance}
\usepackage{xcolor}

\usepackage{stfloats}
\usepackage{siunitx}
\usepackage{url}

\usepackage{subfigure}
\usepackage{comment}

\ifCLASSINFOpdf
\else
\fi
\begin{document}
%


\title{Visual Content Detection in Educational Videos with Transfer Learning and Dataset Enrichment}


\author{
\IEEEauthorblockN{Dipayan Biswas \hspace*{0.2in} Shishir Shah \hspace*{0.2in} Jaspal Subhlok}

\IEEEauthorblockA{Department of Computer Science, 
University of Houston, Houston, USA\\}
\IEEEauthorblockA{ Email: dbiswas@uh.edu, sshah@central.uh.edu, jaspal@uh.edu}
}

\maketitle
\pagestyle{empty} 



%
\IEEEpeerreviewmaketitle

\begin{abstract}

Video is transforming education with online courses and recorded lectures supplementing and replacing classroom teaching. Recent research has focused on enhancing information retrieval for video lectures with advanced navigation, searchability, summarization, as well as question answering chatbots.
Visual elements like tables, charts, and illustrations are central to comprehension, retention, and data presentation in lecture videos, yet their full potential for improving access to video content remains underutilized.
A major factor is that accurate automatic detection of visual elements in a lecture video is challenging; reasons include i) most visual elements, such as charts, graphs, tables, and illustrations, are artificially created and lack any standard structure, and
ii) coherent visual objects may lack clear boundaries and may be composed of connected text and visual components.
Despite advancements in deep learning–based object detection, current models do not yield satisfactory performance due to the unique nature of visual content in lectures and scarcity of annotated datasets. This paper reports on a transfer learning approach for detecting visual elements in lecture video frames. A suite of state-of-the-art object detection models were evaluated for their performance on lecture video datasets. YOLO emerged as the most promising model for this task. Subsequently YOLO was optimized for lecture video object detection with training on multiple benchmark datasets and deploying a semi-supervised auto-labeling strategy. Results evaluate the success of this approach, also in developing a general solution to the problem of object detection in lecture videos. Paper contributions include a publicly released benchmark of annotated lecture video frames, along with the source code to facilitate future research\footnote{\url{https://github.com/dipayan1109033/edu-video-visual-detection}}\footnote{\label{dataset_link}\url{https://github.com/dipayan1109033/LVVO_dataset}}.

\end{abstract}

\section{Introduction}

Lecture videos are widely used as the central component of online education and to enhance classroom learning. Platforms like MOOCs and recorded university lectures offer flexible, self-paced access to diverse subjects \cite{guo2014video, hew2014students, leciaBarker:studentPerceptions}. Slide-assisted lectures, widely preferred by educators and students, promote multimodal learning by combining visuals, text, and narration to reinforce concepts and memory retention \cite{savoy2009information, susskind2005powerpoint, garner2013design}.
Studies in multimedia learning show that integrating visual content such as tables, charts, illustrations along with textual content greatly improves engagement, comprehension, and cognitive processing \cite{mayer2002aids, mayer1991animations, moreno1999cognitive}. In recent years several research projects have focused on improving access to lecture video content, including navigation in long videos \cite{yadav2016vizig}, multimodal summarization \cite{rahman2023enhancing, kawamura2024fastperson}, content adaptation for mobile devices \cite{kim2022fitvid}, and accessibility for visually impaired audiences \cite{peng2021sayitall}. Accurate extraction of visual elements is crucial for all these projects that aim to enhance the value of lecture videos.

This paper focuses on detection of visual objects  in lecture videos.
While object detection in real-world images has been studied extensively, detecting artificially created visual content in lecture video frames, that include charts, graphs and illustrations, presents unique challenges. Visual elements cannot be accurately detected by a traditional approach based on pixel intensity changes
with a sliding window that assumes that visual objects are separated by blank space;  lecture video visual objects are defined by semantic meaning and instructor-driven contexts with varying granularity.

\begin{figure*}[tb]
  \centering
  \subfigure[]{
  \fbox{\includegraphics[width=0.3\textwidth, height=1.8in]{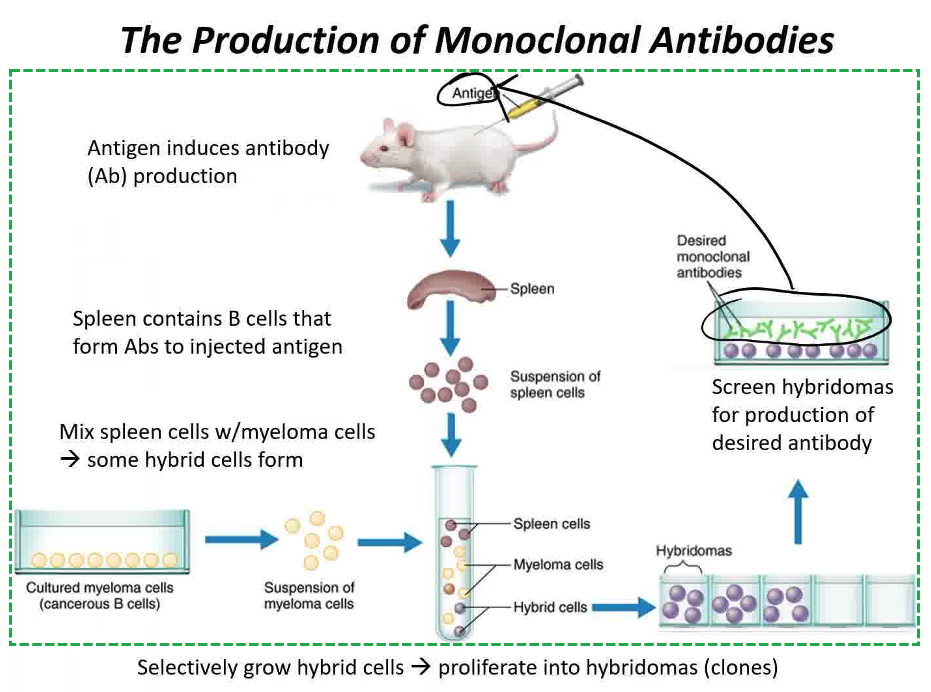}}
  \label{fig:example-slideA}}
  \hfil
  \subfigure[]{
  \fbox{\includegraphics[width=0.3\textwidth, height=1.8in]{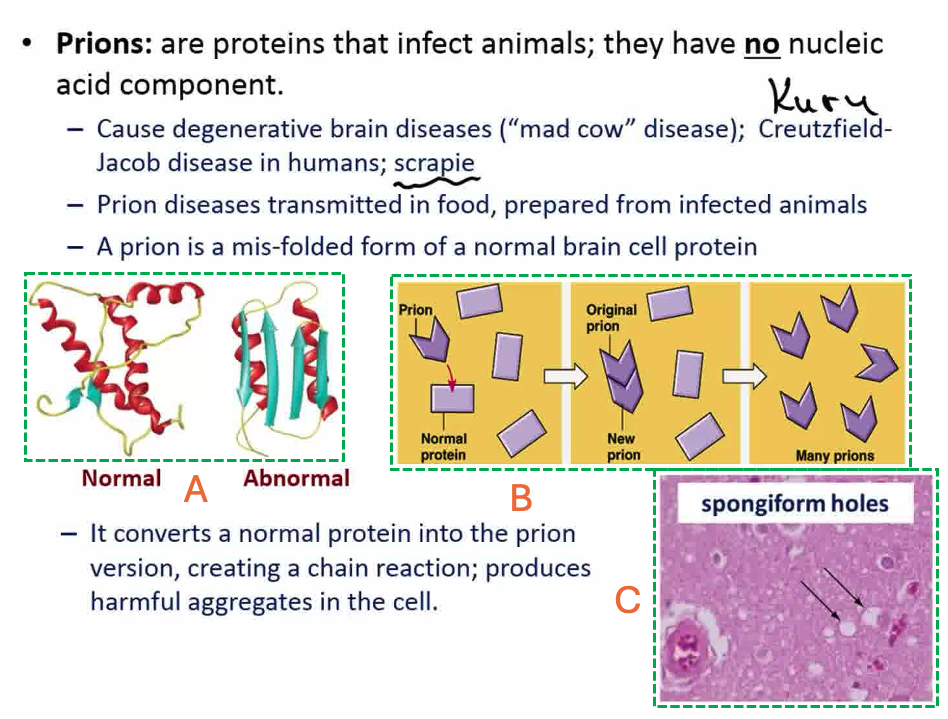}}
  \label{fig:example-slideB}}
  \hfil
  \subfigure[]{
  \fbox{\includegraphics[width=0.3\textwidth, height=1.8in]{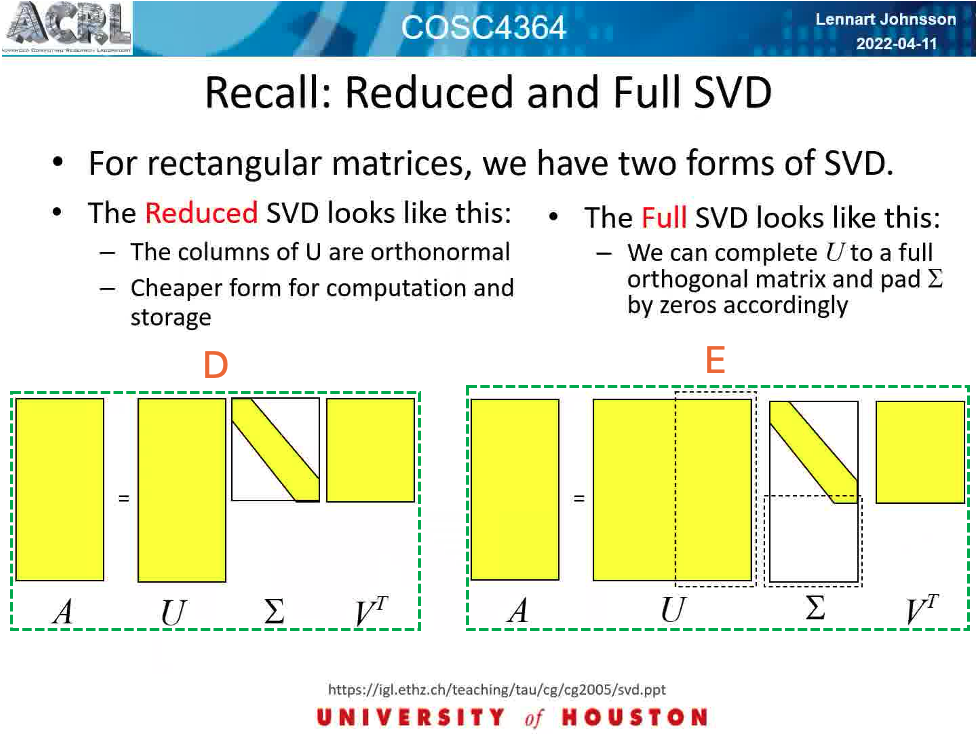}}
  \label{fig:example-slideC}}
  
  \caption{Sample lecture video frames with conceptual visual objects bounded by dotted green rectangles}
  \label{fig:example-slides}
\end{figure*}

We present some examples for illustration. Figure~\ref{fig:example-slideA} shows a lecture video frame containing diverse visual entities that are not individually meaningful but collectively represent a procedural concept. Figure~\ref{fig:example-slideB} is a frame consisting of blocks of visual content in close proximity; they are labeled as three distinct objects (marked A, B, C) based on visual similarity and content semantics, not just physical separation.
Figure~\ref{fig:example-slideC} presents adjacent  components with similar visual appearance that semantically represent two distinct objects, D and E. These examples highlight the complexity of interpreting visual-semantic relationships in lecture slides, where object boundaries and meanings are highly context-dependent, making accurate detection particularly challenging.

Approaches relying on traditional image processing and heuristic algorithms \cite{xu2019lecture2note, biswas2023identifyobject} struggle with complex semantics and space-separated entities, leading to limited success on lecture videos. Deep learning approaches \cite{xu2022semantic, kim2022fitvid, kawamura2024fastperson} show promise, but their effectiveness is limited by scarce annotated datasets and lack of standardization in labeling.

The focus of this paper is accurate identification of visual objects in lecture videos, including chart, graphs and illustrations. The key contributions are the following:
\begin{itemize}
    \item Systematic evaluation of general state-of-the-art object detection models on their effectiveness in identifying visual objects in lecture video frames.
    \item Development and optimization of a transfer learning based approach that employs YOLO and three public lecture video frame datasets for training and evaluation.

    \item Development and evaluation of a semi-supervised pipeline that incorporates dataset enrichment with unlabeled data to enhance the performance of lecture video visual content detection.

    \item Curation and release of a new dataset with 4000 lecture video frames, including 1000 frames with annotated visual objects,
    selected from the collection of lecture videos on Videopoints~\cite{Rahman2023}.
\end{itemize}

This paper is organized as follows: Section 2 reviews relevant related research. Section 3 introduces the datasets used in this study. 
Section 4 compares various object detection models in the context of lecture videos. Section 5 focuses on optimizing and evaluating YOLO for visual object detection. 
Finally, Section 6 states the key findings and conclusions.

\section{Related Work}

Several studies have explored extracting visual elements from presentation slides using metadata \cite{peng2021sayitall, tsujimura2017automaticExpSpot}, but most educational videos are not accompanied by corresponding slides. As a result, pixel-based approaches that analyze video frames directly offer a more general solution. These methods can be grouped into traditional image processing and deep learning techniques. Traditional methods  typically use edge detection and heuristics; for example, Lecture2Note \cite{xu2019lecture2note} merges nearby visual components based on bounding box height and centroid alignment. Recent work on Videopoints \cite{biswas2023identifyobject} employs hierarchical clustering of segmented visual entities, considering spatial proximity, color similarity, and keypoint matching. However, such heuristics often struggle with complex visuals like background changes and animations.

Other research has explored machine learning-based visual element analysis in lecture video frames. Vizig \cite{yadav2016vizig} helps navigation by assigning reference tags to individual slides containing visual elements like charts, diagrams, and tables. This involves classifying slides into six categories using the AlexNet deep learning model trained on unconstrained internet images. 
Shoko et al. \cite{tsujimura2017automaticExpSpot}  classify slide images using SVM trained with downloaded internet images. 
We focus on locating visual elements not clasification.  Kawamura et al. \cite{kawamura2024fastperson} employ ResNet50-FPN based Faster-RCNN object detection model to extract class labels, which are then passed along with textual data to a large-language model to generate summaries. 
FitVid \cite{kim2022fitvid} evaluated several deep learning–based object detectors to extract document design elements for adaptive lecture videos, optimizing readability and customization on small screens. In \cite{xu2022semantic}, the authors introduced a custom deep learning model leveraging ResNet50 with an FPN module to detect both visual and textual elements. 
While these studies developed object detection models using various dataset creation strategies, they lacked a publicly available benchmark for evaluation. Our study focuses on comprehensive evaluation across three different datasets to assess model performance with publicly accessible data. Unlike some prior work that aimed to detect all slide layout elements, we specifically target the detection of image-like visual elements and address the challenge of limited labeled data availability.

\section{Lecture Video Datasets}
\label{sec:datasets}

This study employs three lecture video datasets, primarily consisting of lecture video frames with bounding boxes identifying visual objects, for model training and evaluation.
This research is limited to identification of visual objects, and hence does not utilize category labels that may be available with datasets.

We introduce the {\bf Lecture Video Visual Objects (LVVO)} dataset, constructed from the Videopoints platform \cite{videopoints}, which hosts a large archive of screen-captured live lectures. The dataset comprises 4,000 video frames extracted from 245 lecture videos spanning 13 courses in biology, computer science, and geosciences, recorded between 2019 and 2024. Frames were selected based on visual richness and filtered to remove near-duplicates ensuring diversity.
A randomly selected subset of 1,000 frames (\texttt{LVVO\_1k}) was manually annotated with bounding boxes and one of four visual categories. Each frame was independently labeled by two expert annotators, with any discrepancies reviewed and resolved by a third expert. The remaining 3,000 frames (\texttt{LVVO\_3k}) were annotated using a semi-supervised auto-labeling approach discussed in Section~\ref{sec:auto-labeling}. This multi-stage process ensured high-quality annotations suitable for benchmarking object detection in lecture videos.

We also employed two external datasets: the {\bf Lecture Design Dataset (LDD)} \cite{kim2022fitvid}, created to support layout adaptation and readability in online courses, and the {\bf Lecture Presentations Multimodal (LPM)} dataset \cite{lee2023lectureMDataset}, intended for training vision-language models to interpret and describe slide visuals. Both datasets include bounding box annotations that were 
employed in this research. 
To ensure consistency, we filtered all datasets by removing frames without visual elements, near-duplicates, purely textual content, and objects with widely separated textual components lacking visual coherence. Key elements of the filtered versions of the external and the LVVO dataset used in our experiments are
summarized in Table~\ref{tab:dataset_comparison}.
Details of the new LVVO dataset, along with the filtered versions of external datasets are publicly available\textsuperscript{\ref{dataset_link}} \cite{biswas2025lvvodataset}.

\begin{table}[htb]
    \centering
    \renewcommand{\arraystretch}{1.2} 
    \caption{Key attributes of datasets used in this research}
    \resizebox{\columnwidth}{!}{ 
    \begin{tabular}{|l|c|c|c|c|}
        \hline
        \multirow{2}{*}{Attribute} & \multirow{2}{*}{LDD dataset} & \multirow{2}{*}{LPM dataset} & \multicolumn{2}{c|}{LVVO dataset} \\
        \cline{4-5}
        & & & LVVO\_1k & LVVO\_3k \\
        \hline
        \#Images & 4387 & 3981 & 1000 & 3000 \\
        \#Objects & 7052 & 5688 & 1580 & 5073 \\
        Objects/Image & 1.61 & 1.43 & 1.58 & 1.69 \\
        \hline
        Content type & online course & youtube video & \multicolumn{2}{c|}{recorded lecture} \\
        \#Courses & 66 & 35 & \multicolumn{2}{c|}{13} \\
        \#Videos & NA & 334 & \multicolumn{2}{c|}{245} \\
        \hline
        \multicolumn{5}{|c|}{Object Count Distribution} \\
        \hline
        0-1 objects & 61.59\%  & 71.19\%  & 63.90\%  & 58.07\% \\
        2-3 objects & 33.30\%  & 25.90\%  & 30.70\%  & 34.83\% \\
        $>=$ 4 objects & 5.11\%  & 2.91\%  & 5.40\%  & 7.10\% \\
        \hline
    \end{tabular}
    }
    \label{tab:dataset_comparison}
\end{table}

\section{Performance of Object Detection Models on Lecture Video Datasets}
\label{sec:object_detection_models}

This section presents the development of object detection models for lecture video frames, leveraging transfer learning with limited labeled datasets. 
We assess detection performance using widely adopted COCO metrics \cite{lin2014microsoftCOCO} (AP, AP50, AP75) based on Intersection over Union (IoU). AP50 and AP75 represent lenient (0.50) and strict (0.75) thresholds, respectively, while overall AP is the average precision computed across multiple IoU thresholds (0.50 to 0.95). We also report Precision, Recall, and F1 Score where relevant.

\textbf{Models and Training:}
We evaluate six state-of-the-art general object detection models: Faster R-CNN \cite{ren2016faster}, SSD \cite{liu2016ssd}, Mask R-CNN \cite{he2017mask}, RetinaNet \cite{lin2017focalRetinaNet}, FCOS \cite{tian2022fullyFCOS}, and YOLOv11 \cite{ultralytics2024yolo11docs}. These models, pre-trained on the COCO dataset \cite{lin2014microsoftCOCO}, were fine-tuned using transfer learning. This approach is particularly effective when labeled data is limited, as it enables the models to leverage transferable features learned from a large source dataset.
As demonstrated by Yosinski et al. \cite{yosinski2014transferable}, the lower layers of deep neural networks capture generalizable features such as edges and textures that remain effective across different domains, while the higher layers learn more task-specific patterns that require adaptation during fine-tuning.

We fine-tuned each model by freezing the early convolutional layers to preserve general features while updating the higher layers using a limited set of annotated lecture video frames.
All models are fine-tuned on three datasets (LVVO\_1k, LDD, and LPM) with a consistent 80\%-20\% train-validation split to ensure reliable and comparable evaluation. Pre-trained weights from the COCO dataset are used for initialization via the \texttt{ultralytics} framework and PyTorch’s \texttt{torchvision} library. For YOLOv11, we adopt the medium variant and freeze the first three blocks; for remaining ResNet50-based models, the first two convolutional blocks are frozen to preserve low-level transferable features. All models are trained under a unified configuration—learning rate of 0.001, batch size of 8, and 30 epochs. These are selected empirically  based on best practices and performance across experiments.

\begin{figure}[tb]
    \centering
    \includegraphics[width=0.8\linewidth]{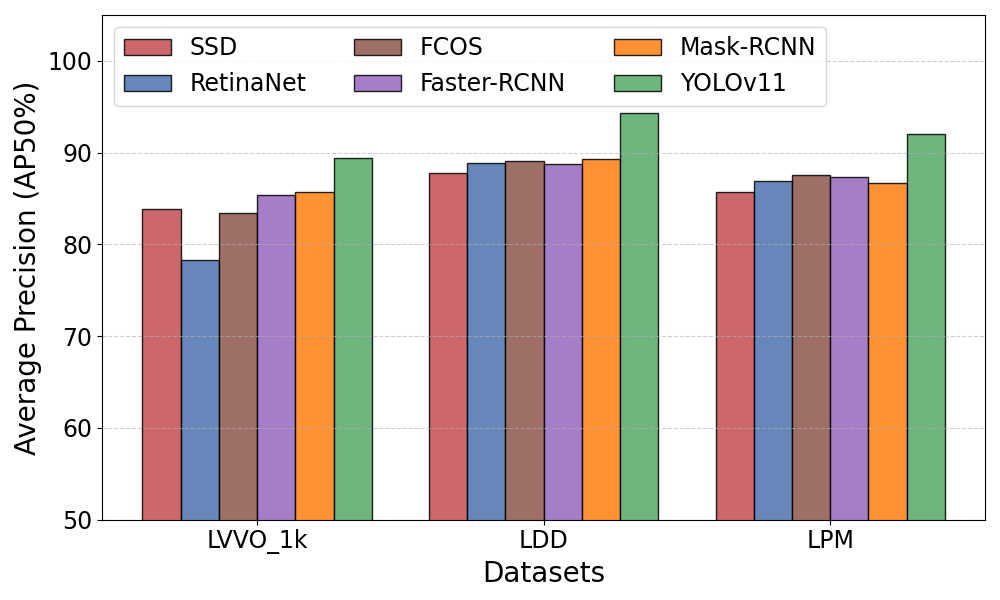}
    \caption{Average Precision (AP50\%) comparison of different object detection models on three datasets: \textit{LVVO\_1k}, \textit{LDD}, and \textit{LPM}. 
    }
    \label{fig:models-comparison}
\end{figure}

\textbf{Results:} 
Figure~\ref{fig:models-comparison} presents the experimental results comparing object detection models across three datasets: \textit{LVVO\_1k}, \textit{LDD}, and \textit{LPM}. Key observations are:

\begin{itemize}
    \item Highest performance achieved exceeds 90\% for AP50\%, which is encouraging for this challenging problem.
    \item YOLOv11 consistently outperformed other models across all datasets.
    \item Best performance is achieved for {\em LDD} dataset with {\em LPM} dataset showing a similar pattern with a slightly lower performance. 
    For  \textit{LVVO\_1k} (1000 images) dataset the performance is lower, especially for RetinaNet, suggesting limited robustness in low-data scenarios~\cite{li2020metaRetinaNet}.
\end{itemize}

Because of consistent superior performance, YOLOv11 was selected for further experimentation in this study.

\begin{figure}[bt]
    \centering
    \includegraphics[width=0.7\linewidth]{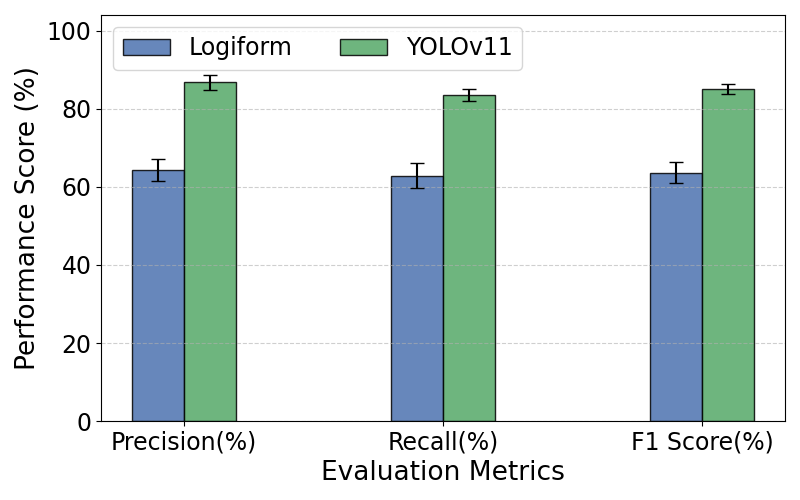}
    \caption{Performance comparison between the \textit{Logiform} algorithm \cite{biswas2023identifyobject} and YOLOv11 on the \textit{LVVO\_1k} dataset using 5-fold cross-validation.
    }
    \label{fig:logiform_yolo11}
\end{figure}

We evaluated the relative performance of \textit{Logiform} \cite{biswas2023identifyobject}, an advanced pixel analysis based algorithm  that employs hierarchical clustering on segmented visual entities to capture semantic coherence, leveraging spatial proximity, color similarity, and keypoint matching. 
Figure~\ref{fig:logiform_yolo11} compares the performance of the \textit{Logiform} algorithm with  YOLO11 model on the \textit{LVVO\_1k} dataset. YOLO11 significantly outperforms \textit{Logiform}, highlighting the power of deep learning-based models for identifying visual content in lecture video frames.

\section{Optimizing YOLO for Lecture Videos}
In this section, we focus on optimizing the top performing YOLO model for accurately identifying visual objects in lecture videos. 
YOLOv11's architectural advancements such as the C3k2 block, SPPF module, and C2PSA module contribute significantly to its superior performance \cite{ultralytics2024yolo11docs}. The C3k2  (Cross Stage Partial with kernel size 2) block improves feature aggregation, enhancing efficiency and speed. The SPPF (Spatial Pyramid Pooling - Fast) module facilitates multi-scale detection, enabling accurate recognition of objects of varying sizes. Additionally, the C2PSA (Convolutional Block with Parallel Spatial Attention) module enhances spatial attention, allowing a focus on critical regions within lecture frames.

\subsection{Cross-dataset Evaluation}
\begin{figure}[tb]
    \centering
    \includegraphics[width=0.99\linewidth]{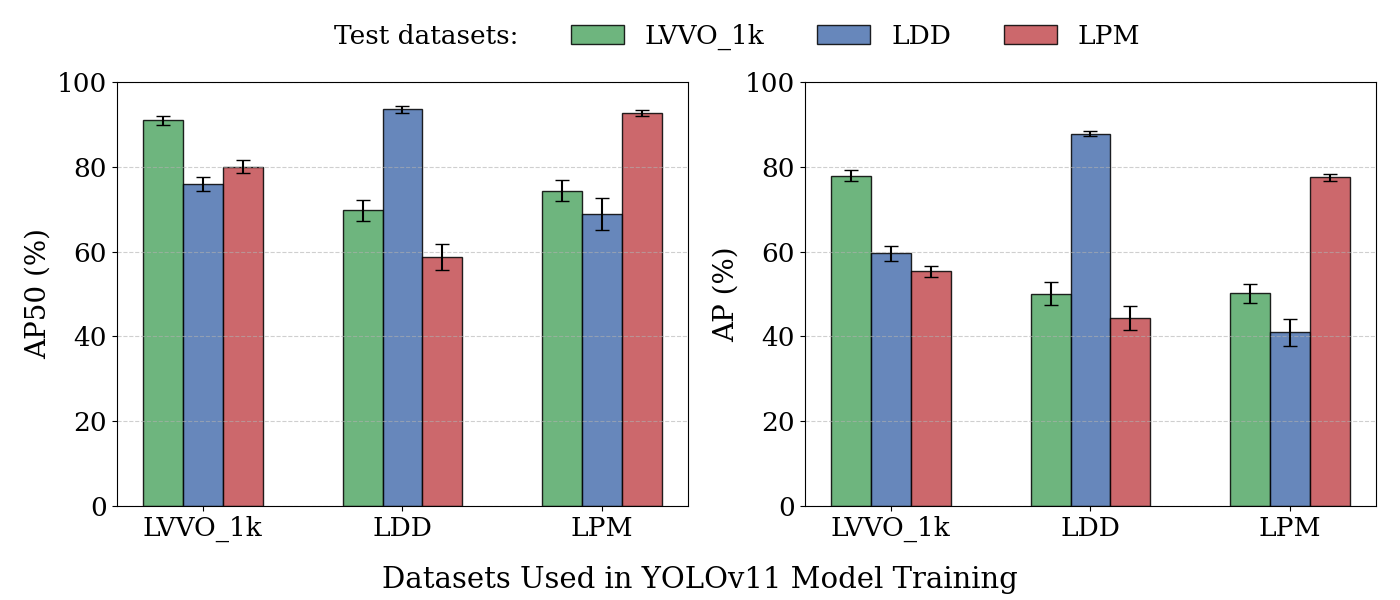}
    \caption{Performance of object detection models trained on one dataset (labels in x-axis) and tested on other datasets with AP50 (left) and Overall AP (right) metrics. The green, blue, and red bars represent \textit{LVVO\_1k}, \textit{LDD}, and \textit{LPM} dataset, respectively.}
    \label{fig:cross_dataset_testing}
\end{figure}

This research has employed 3 datasets discussed in section \ref{sec:datasets}.
We conducted experiments to assess how models trained on one lecture video dataset perform on detection tasks on other datasets in the same domain.
The results are presented in Figure~\ref{fig:cross_dataset_testing}. All  models perform significantly worse when entirely trained on other datasets, as compared to the case when a part of the same dataset is employed for training. 
Overall AP (right) drops more than AP50 (left) due to its stricter localization criteria that penalize imprecise bounding boxes.

These results highlight the challenge of developing an accurate model for diverse lecture videos. 
While all datasets employed contain labeled visual objects in lecture video frames, there are significant differences. Datasets are dominated by different disciplines as well as diverse artificial slide formats. Other characteristics that vary across datasets include i) presence and location of instructor faces, ii) presentation and recording software employed, and iii) varying quiz platforms and visual teaching tools.  These factors underscore the need for dataset-specific adaptation.

\subsection{Building a Generalized Model}

The focus of this section is developing a general model that is accurate across datasets.
To achieve this, we fine-tune the COCO YOLOv11 \cite{yolo11_ultralytics} model on a combined dataset consisting of images from \textit{LDD}, \textit{LPM}, and \textit{LVVO\_1k} datasets. Each dataset is split into a 70\%:10\%:20\% ratio for training, validation, and testing. The model is fine-tuned using the combined training splits and validation splits. After training, we evaluate the model separately on each individual dataset's test split. Three scenarios are evaluated:
i) {\em Full Training}: All images in all data sets are employed for training subject to the 70\%:10\%:20\% split ii) {\em Limited Training:} 1000 images are employed for training in each dataset, and iii) {\em Baseline:} Only the dataset used for training is included for testing.

\begin{figure}[tb]
    \centering
    \includegraphics[width=0.99\linewidth]{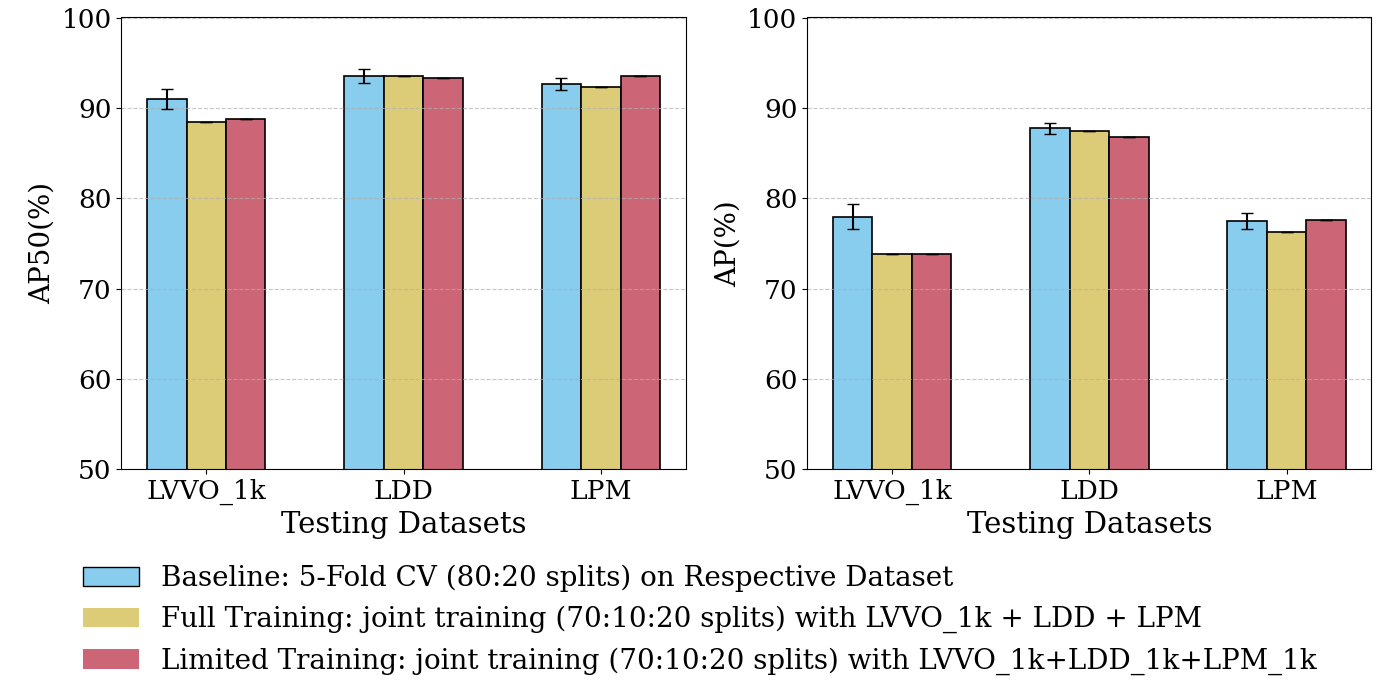}
    \caption{Comparison of two joint training strategies across three datasets with the "baselines" (light blue), defined as 5-fold cross-validation performance on individual datasets. Yellow bars show joint training with all images ($\sim$4000 samples) from \textit{LDD} and \textit{LPM} datasets, while red bars uses 1000 sample images per dataset. Metrics shown: AP50 (left) and Overall AP (right).}
    \label{fig:joint_training}
\end{figure}

The results are presented in Figure~\ref{fig:joint_training}.
The key result is that, for each dataset, a jointly trained model's performance is comparable to a model trained exclusively on the test dataset. This demonstrates that models can learn unique aspects of datasets without diluting the knowledge gained when training is limited to one dataset.
Furthermore, there is minimal difference in performance when training is limited to 1000 images versus all images (up to 4000).
This suggests that a modest, diverse set of images capturing distinct visual characteristics and content variations can support generalization. 

\begin{figure}[bt]
    \centering
    \includegraphics[width=0.8\linewidth]{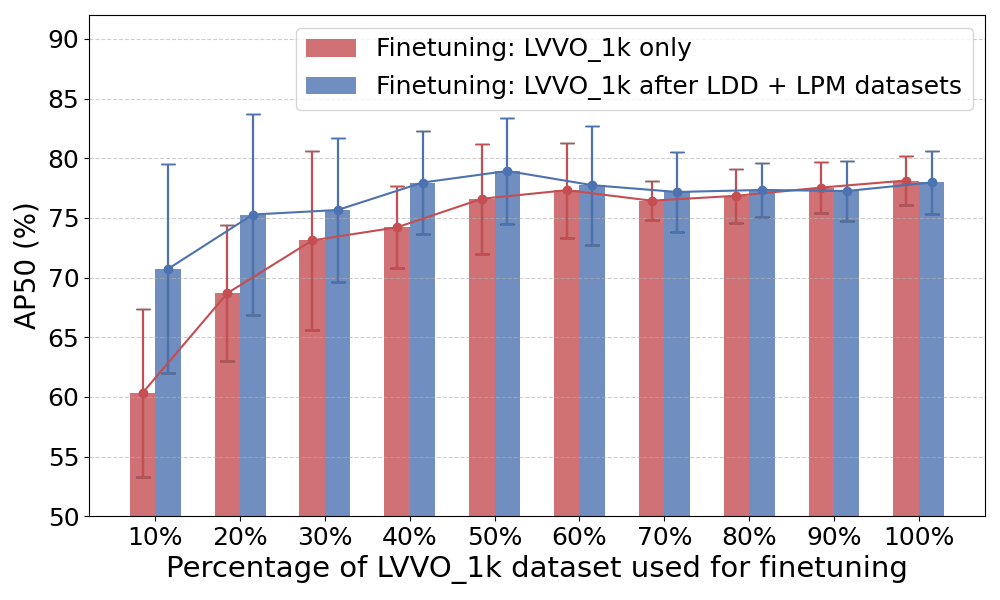}
    \caption{Impact of additional finetuning with LDD and LPM datasets on performance for LVVO\_1k dataset.} 
    
    \label{fig:cross_datasets_pretrain}
\end{figure}

When a labeled dataset is small, developing an accurate model is challenging, and other in domain datasets can be helpful for training. We conducted a set of experiments to investigate this relationship.
Specifically, we compare two scenarios: COCO YOLOv11 model \cite{yolo11_ultralytics} (1) fine-tuned directly on the target dataset (\textit{LVVO\_1k}) and (2) additional training on a larger in-domain dataset (\textit{LDD} + \textit{LPM}) before fine-tuning on \textit{LVVO\_1k}.
Figure~\ref{fig:cross_datasets_pretrain} compares performance in  both scenarios across varying target dataset sizes. 
We note that leveraging LDD and LPM datasets is beneficial for a model for LVVO\_1k when labeled data is scarce (e.g 10\%-20\% of 1K) but there is no significant benefit when labeled data is above a threshold (around 60\% of 1K). Hence training with external datasets is likely to be important in this domain only for small datasets.

\subsection{Dataset Enrichment} 
\label{sec:auto-labeling}

Manual annotation of lecture video data
is resource-intensive and time-consuming. 
In contrast, data without manual annotation is relatively easy to assemble. For the LVVO dataset, we have 3000 frames without annotation, in addition to 1000 annotated frames. We explore if this unlabeled data can be helpful in improving the accuracy of visual object detection.

Our methodology involves training a deep learning model on a smaller manually labeled dataset and using it to generate additional annotations for unlabeled images. Specifically, we fine-tune a COCO-pretrained YOLOv11 model \cite{yolo11_ultralytics} using transfer learning, first adapting it to the manually annotated \texttt{LVVO\_1k} dataset, which was divided into 80\% for training and 20\% for validation. The fine-tuned model is then employed in inference mode to predict bounding boxes for the unlabeled images, with a confidence threshold of 0.5 applied to filter out low-confidence detections. This process results in an automatically labeled dataset, \texttt{LVVO\_3k}, which effectively expands the \textit{LVVO} dataset to 4,000 annotated images.

\begin{figure}[tb]
    \centering
    \includegraphics[width=0.7\linewidth]{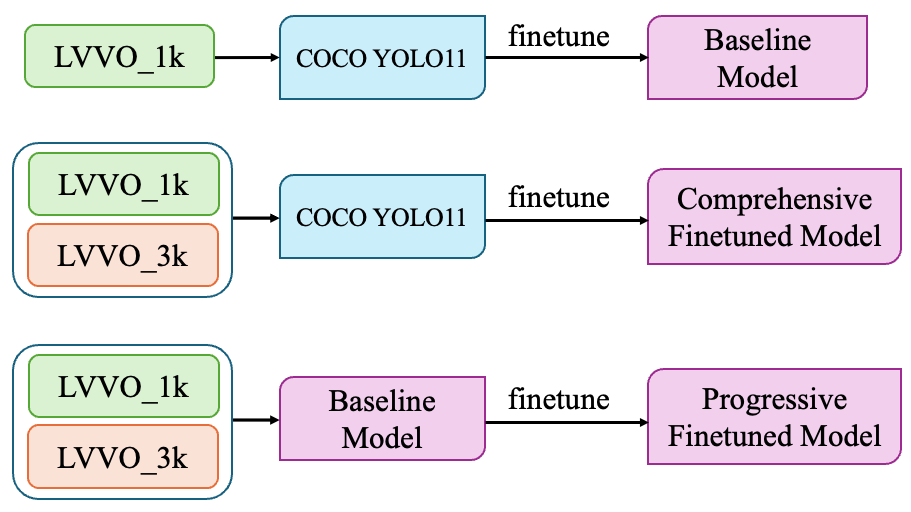}
    \caption{Different fine-tuning approaches using the extended dataset. Where, \textit{COCO YOLOv11} \cite{yolo11_ultralytics} is the pretrained YOLOv11 model.}
    \label{fig:finetune_extended}
\end{figure}

Figure~\ref{fig:finetune_extended} illustrates two primary fine-tuning strategies that leverage the extended labeled dataset to improve model performance.
For comparison, we first finetune the COCO-pretrained YOLOv11 model \cite{yolo11_ultralytics} solely on our manually labeled \textit{LVVO\_1k} to establish a {\em Baseline} model. Standard 5-fold cross-validation was applied for this baseline and for the following two fine-tuning approaches:  

\begin{itemize}
    \item {\em Comprehensive Fine-Tuning:}  We fine-tune the COCO-pretrained YOLOv11 model using the entire extended dataset: 1k manually labeled + 3k auto-labeled images.  
    \item {\em Progressive Fine-Tuning:} We start with the Baseline model optimized on the 1k manually labeled images, and continue fine-tuning with the entire extended dataset.  
\end{itemize}

Table~\ref{tab:training-approach} summarizes the performance results on the same cross-validation images of manually labeled \textit{LVVO\_1k} dataset. Progressive Finetuning yields the highest improvement, increasing AP50 from 90.75\% to 95.32\% and AP from 77.6\% 84.19\%, a 6.59 percentage point increase over the Baseline. These findings indicate that leveraging the auto-labeled dataset substantially enhances model performance.
Furthermore, Progressive Finetuning slightly surpasses Comprehensive Finetuning, indicating that building a strong initial foundation followed by gradually refining with diverse examples enhances model adaptability.

\begin{table}[bt]
    \centering
    \renewcommand{\arraystretch}{1.3} 
    \setlength{\tabcolsep}{2pt} 
    \caption{Impact of auto-labeling on model performance across different fine-tuning strategies.}
    \label{tab:training-approach}
    \begin{tabular}{lcccc}
        \hline
        \textbf{Model} & \textbf{AP50 (\%)} & \textbf{AP75 (\%)} & \textbf{AP (\%)} & \textbf{F1 Score (\%)} \\
        \hline
        Baseline & 90.75 & 83.91 & 77.6 & 85.14 \\
        Comprehensive Finetuned & 94.67 & 90.15 & 83.89 & 89.44 \\
        Progressive Finetuned & \textbf{95.32} & \textbf{90.48} & \textbf{84.19} & \textbf{89.93} \\
        \hline
    \end{tabular}

\end{table}

\begin{figure}[bt]
    \centering
    \includegraphics[width=0.8\linewidth]{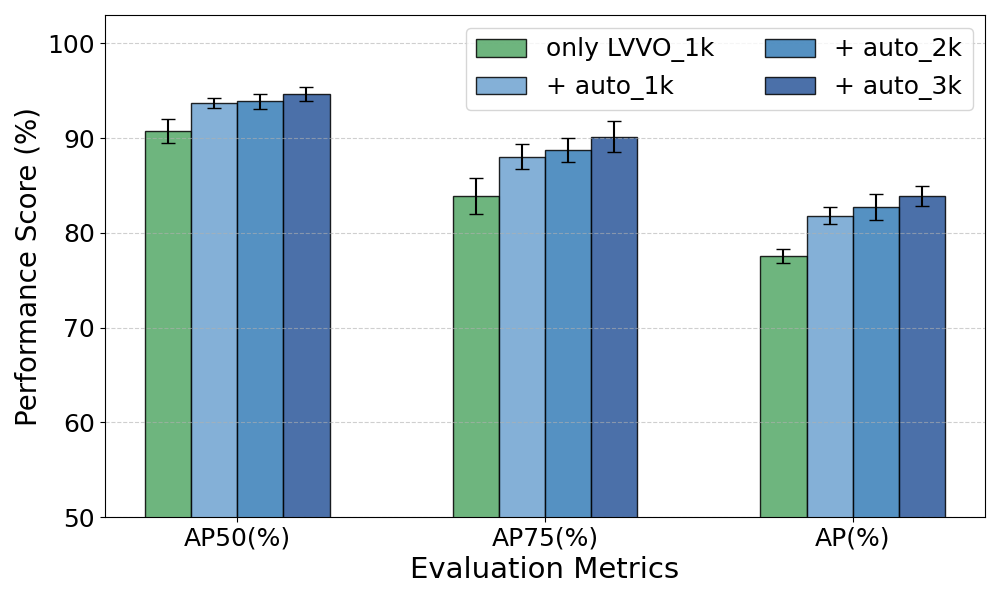}
    \caption{Performance improvement across AP50, AP75, and AP metrics as 1k, 2k, and 3k auto-labeled images are incorporated during fine-tuning. 
    }
    
    \label{fig:incremental_autolabel_images}
\end{figure}

Figure~\ref{fig:incremental_autolabel_images} illustrates the model's performance when 1k, 2k, and 3k auto-labeled images were added to the manually labeled dataset.
We observe a significant performance boost with the addition of 1k auto-labeled images, while subsequent increments yield smaller improvements. Performance improvement is more pronounced for stringent metrics such as AP75 and AP. The conclusion is that incorporating auto-labeled images can meaningfully improve visual content detection for a small labeled training data set.

\section{Conclusions}

The key contribution of this paper is developing transfer learning based models to accurately detect visual objects in lecture video frames for limited labeled dataset scenario. The research was conducted with three lecture video frame datasets including Lecture Video Visual Objects (LVVO), a new dataset created as part of this research and available to the research community.

We evaluated six general state-of-the-art object detection models for their effectiveness in detecting visual objects in lecture video domain with fine tuning. YOLOv11 emerged as the best model. We then investigated several approaches to further improve the accuracy of YOLOv11 for this task.

We observed that models exclusively trained on one dataset exhibited considerable performance degradation when tested on other datasets. This is not surprising since each dataset has some unique visual characteristics, but it highlights the challenge of building a general object detection model. We were able to build a general model with training on parts of all datasets which was almost as accurate as a model that was only trained on a specific dataset. This general model achieved AP50 performance between 88.8\% and 93.5\% depending on the three datasets.
This is an important result as it demonstrates an effective approach to building a general purpose object detector for lecture videos. 

A major challenge in applying deep learning to a new problem is the limited availability of labeled datasets. For the LVVO dataset, we have 1000 labeled frames, and another 3000 unlabeled frames. We explored a data enrichment approach  to exploit unlabeled frames for improving performance. The unlabeled data was automatically labeled, and then integrated into training. This led to an increase in AP50 performance from 90.7\% to 95.3\%. The results also demonstrate the relationship between the size of the auto-labeled dataset and performance improvement.

Overall, our study provides valuable insights into optimizing object detection for lecture video analysis, reinforcing the importance of dataset enrichment and adaptation strategies. Future research will explore  adaptive fine-tuning techniques and semi-supervised learning approaches while minimizing the dependency on manual annotations.

\section*{Acknowledgement}
The authors wish to express their sincere gratitude to all current and former members of the Videopoints team, especially  Jatindera  Walia who manages the frameworks employed. Partial support was received from the National Science Foundation under award NSF-SBIR-1820045. This work was completed in part with resources provided by the Research Computing Data Core at the University of Houston.

\bibliographystyle{IEEEtran}


\end{document}